\documentclass[11pt]{article}

\usepackage{acl}

\usepackage{times}
\usepackage{latexsym}
\usepackage{float}

\usepackage[T1]{fontenc}

\usepackage[utf8]{inputenc}

\usepackage{microtype}

\usepackage{inconsolata}

\usepackage{graphicx}

\usepackage{xcolor}
\usepackage{xcolor,colortbl}
\definecolor{hidden-draw}{RGB}{0,0,0}

\usepackage{tikz}
\usepackage[edges]{forest}
\usepackage[most]{tcolorbox}

\usetikzlibrary{trees,positioning,shapes,shadows,arrows.meta}

\usepackage{amsmath, amssymb, amsfonts}

\usepackage{placeins}
\usepackage{booktabs} 
\usepackage{multirow}
\usepackage{pgfplots}

\usepackage{array}

\pgfplotsset{compat=1.18}
\title{LLM as Attention-Informed NTM and Topic Modeling as long-input Generation: Interpretability and long-Context Capability}

\author{
     Xuan Xu\textsuperscript{1,*} 
     \ Zhongliang Yang\textsuperscript{1,*} 
     \ Haolun Li\textsuperscript{1,*} 
     \ Beilin Chu\textsuperscript{1} \\ 
     \ \textbf{Rui Tian}\textsuperscript{2}
     \ \textbf{Yu Li}\textsuperscript{3} 
     \ \textbf{Shaolin Tan}\textsuperscript{4} 
     \ \textbf{Linna Zhou}\textsuperscript{1,$\dagger$} \\
    \textsuperscript{\rm 1} Beijing University of Posts and Telecommunications \\
    \textsuperscript{\rm 2} University of Science and Technology Beijing \\
    \textsuperscript{\rm 3} Beijing Value Simplex Technology Co. Ltd \\
    \textsuperscript{\rm 4} Zhongguancun Laboratory \\
    {\tt\{sh22xuxuan,yangzl,lhldudu,beilin.chu,songj,zhoulinna\}@bupt.edu.cn} \\
    {\tt u202343368@xs.ustb.edu.cn \quad
    liyu@entropyreduce.com \quad
    shaolintan@hnu.edu.cn} \quad
}

\begin{document}
\setlength\titlebox{6.5cm}

\maketitle
\renewcommand{\thefootnote}{\fnsymbol{footnote}}
\footnotetext[1]{These authors contributed equally.}
\footnotetext[2]{Corresponding author.}
\renewcommand{\thefootnote}{\arabic{footnote}}

\begin{abstract}
Topic modeling aims to produce interpretable topic representations and topic--document correspondences from corpora, but classical neural topic models (NTMs) remain constrained by limited representation assumptions and semantic abstraction ability. We study LLM-based topic modeling from both white-box and black-box perspectives. For white-box LLMs, we propose an attention-informed framework that recovers interpretable structures analogous to those in NTMs, including document-topic and topic-word distributions. This validates the view that LLM can serve as an attention-informed NTM. For black-box LLMs, we reformulate topic modeling as a structured long-input task and introduce a post-generation signal compensation method based on diversified topic cues and hybrid retrieval. Experiments show that recovered attention structures support effective topic assignment and keyword extraction, while black-box long-context LLMs achieve competitive or stronger performance than other baselines. These findings suggest a connection between LLMs and NTMs and highlight the promise of long-context LLMs for topic modeling.
\end{abstract}

\section{Introduction}
Topic modeling (TM) aims to uncover the latent semantic structure of large-scale corpora. 
Classical probabilistic models such as Latent Dirichlet Allocation (LDA) \cite{blei2003latent,Griffiths2004FindingST} 
represent each document as a topic mixture $\theta_d$ on the simplex and each topic as a word distribution $\phi_k$, 
typically with Dirichlet priors to encourage sparsity and interpretability. 
This explicit ``document--topic / topic--word'' latent-variable framework yields low-dimensional and interpretable structure, 
but usually relies on the bag-of-words (BoW) assumption and simple priors, often leading to topic mixing or low informativeness on complex corpora.

Neural topic models (NTMs) \cite{wuSurveyNeuralTopic2024} extend this framework with neural networks, 
often using VAE-style amortized inference. 
Models such as ProdLDA \cite{srivastava2017autoencoding} and ETM \cite{dieng2020topic} improve inference efficiency and semantic coherence, 
while later work improves topic quality through alignment, disentanglement, contextualized representations, and regularization 
\cite{zhao2020neural,wu-etal-2021-discovering,bianchi-etal-2021-pre,wu-etal-2022-mitigating,pmlr-v202-wu23c}. 
Nevertheless, the paradigm remains largely grounded in BoW or weak contextual signals, 
limiting its ability to model word order, discourse structure, long-range dependencies, 
and external knowledge for semantic abstraction.
With the rise of large language models (LLMs) \cite{yang2024harnessing,xi2025rise}, many studies have begun to use prompting and generation to directly perform topic extraction and corpus organization, forming an engineering pipeline of ``generation--refinement--assignment.''\cite{phamTopicGPTPromptbasedTopic2024a,doiTopicModelingShort2024,bnaic2024_small_data_llm_topic} Empirical evidence suggests that direct prompting for topic discovery and assignment can match or even substitute traditional end-to-end topic models \cite{mu-etal-2024-large}. 
Subsequent work explores prompt design and workflow strategies: PromptTopic targets sentence-level extraction for short texts \cite{wangPromptingLargeLanguage2023}; \citet{spielberger2025retrieval} introduces a retrieval-augmented generation (RAG) framework for topic modeling in organizational research. Extensions such as CHIME \cite{hsuCHIMELLMAssistedHierarchical2024b} and LLooM \cite{lamConceptInductionAnalyzing2024} address hierarchical organization and concept induction , while AgenTopic \cite{AgenTopic} integrates LLMs into iterative refinement and model selection.

In general, these studies highlight the advantages of LLMs in semantic abstraction and instruction following, but most existing work places greater emphasis on system design and engineering implementation, pays less attention to a deeper mechanistic understanding of LLM-based TM, and does not examine the intrinsic connection between LLMs and classical NTM methods. 
In addition, previous methods have not adequately realized that the rapidly growing context windows of modern LLMs \cite{liu2025comprehensive} range from several hundred thousand to several million tokens which could address small corpora at once.




From an information-geometric perspective, LLM can be viewed as an implicit continuous semantic manifold model~\cite{zhang2025multi}, whose representations can be decomposed into three semantic manifolds: local (word-level), intermediate (sentence-level) and global (discourse-level).
Unlike LDA or NTMs, which perform document-level decomposition in a low-dimensional latent space, LLMs model context end-to-end in a high-dimensional continuous space. 
Explicit topic-mixture variables disappear but may instead be embedded in hidden states and attention in a distributed form. 
Previous work has shown that interpretable word groups can be disentangled from neurons or pathways \cite{lim-lauw-2023-disentangling}, suggesting that LLMs can be reinterpreted as NTMs under superposition, where topic structures exist as distributed and overlapping semantic factors.

This paper focuses on the structural existence question corresponding to traditional NTMs 
(\textbf{RQ1:} if LLMs implicitly perform topic modeling, can we recover interpretable document--topic and topic--word structures from internal states?). 
We probe white-box LLMs through multi-level attention aggregation, and empirically show that interpretable NTM-like structures can be recovered from prompt-side and generation-side attention. This leads to a perspective that views LLMs as attention-informed NTMs.

Since white-box access is often unavailable and SOTA black-box LLMs have million-level context windows and can meet the needs of corpus-level topic modeling, we further study a black-box framework 
(\textbf{RQ2:} how to improve applicability under black-box constraints, especially for long-context generalization?). 
We reformulate topic modeling as a structured long-input task similar to tasks in long-form benchmarks such as LOFT \cite{leeCanLongContextLanguage2024a} and  LongInOutBench \cite{zhang2025lost} and propose a ''post-generation signal compensation'' strategy, which approximates latent topic-mixture signals via diversified topic-cue generation and hybrid retrieval in an external memory space. 
This enables stable topic modeling without access to internal mechanisms, and we also evaluate SOTA black-box LLMs under ultra-long contexts compared with NTMs and other advanced TM models. We attempt to demonstrate the potential of current ultra-long-context LLM for topic modeling tasks and provide a reference for the subsequent use of TM as a norm in LLM competency assessments. Our main contributions are as follows.
\begin{itemize}
    \item We establish a formal connection between NTMs and LLMs, proposing the perspective of LLMs as attention-informed NTMs.
    
    \item We design attention-based probing methods to recover interpretable topic structures in white-box LLM-based topic modeling.
    
    \item We propose a black-box framework with post-generation signal compensation for practical LLM API-based topic modeling.
    
    \item We conduct comprehensive evaluations on long-context topic modeling, demonstrating the potential of ultra-long-context LLM and positioning topic modeling as a long-input benchmark for assessing LLM competencies.
\end{itemize}

\begin{figure*}[!t]
\centering
\resizebox{\textwidth}{!}{%
\includegraphics[width=\textwidth]{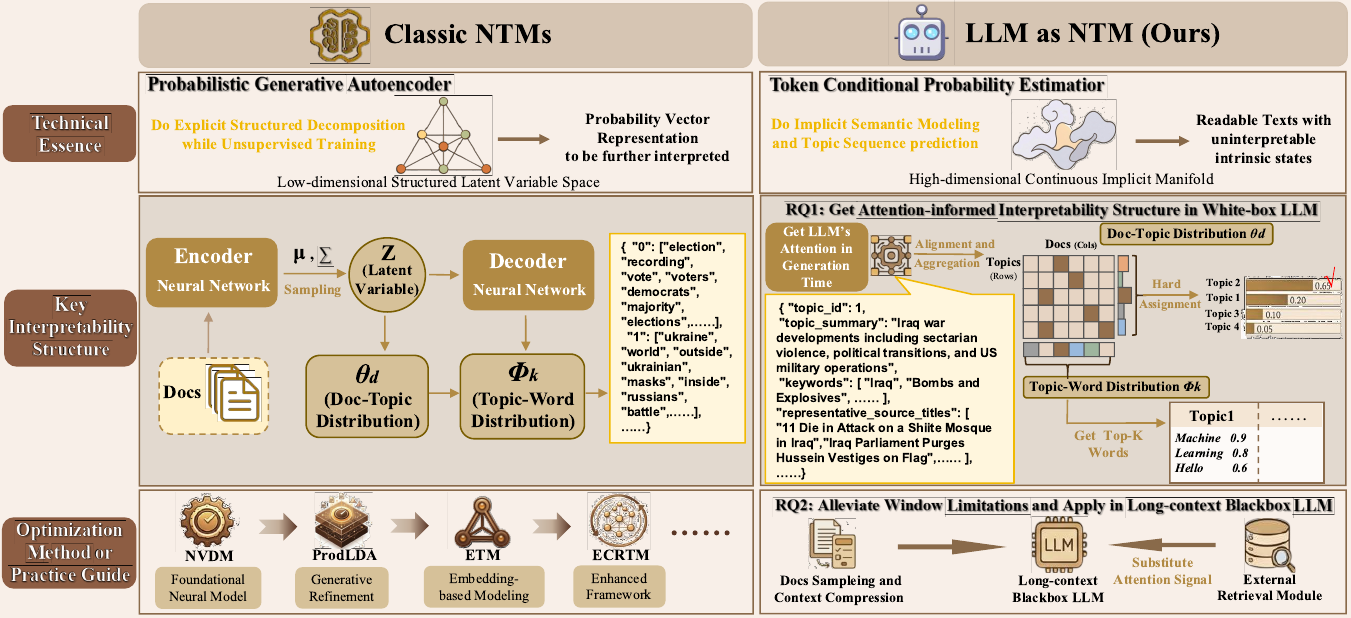}
}
\caption{Theoretical and Methodological Comparison Between Traditional NTMs and Our LLM-as-NTM Framework}
\label{fig1}
\end{figure*}

\section{Background and Motivation}

\subsection{Interpretable Structure of NTMs}

NTMs perform approximate inference on document-level latent variables and reconstruct documents through a decoder. Their core interpretable structure lies in two explicit components: each document is represented by a low-dimensional topic mixture $\theta_d$, and each topic is modeled as a word distribution $\phi_k$. Words are generated from the interaction between $\theta_d$ and $\phi_k$, typically under the bag-of-words assumption, and the training objective (e.g., ELBO) directly optimizes these variables, yielding interpretable document--topic and topic--word correspondences.

\subsection{Limitations of VAE-based Topic Modeling}
Despite their success, VAE-based NTMs implicitly restrict the notion of a ``topic'' to a probability distribution over the vocabulary, with documents represented by mixture vectors $\theta_d$. This formulation emphasizes lexical salience and bag-of-words reconstruction, but struggles to capture richer semantic structures such as events, relations, stances, or hierarchical organization. In modern applications—e.g., short texts, cross-domain corpora, and interactive analysis—users increasingly require structured and controllable topic representations that cannot be naturally reduced to static word distributions. In addition, alignment is mediated primarily through $\theta_d$, which offers limited auditable evidence for why a document belongs to a topic. The LLM-based TM paradigm offers a higher-capacity alternative.

\subsection{A More Fundamental Definition of Topic Modeling}

We detach topic modeling from the specific parameterization of $(\theta,\phi)$ and define it functionally: given a corpus, a topic model outputs human-readable topic representations and testable topic--document correspondences.

Let $\mathcal{C}=\{d_1,\dots,d_N\}$. A topic model is a mapping
\begin{equation}
\mathcal{X}: \mathcal{C} \mapsto (\mathcal{T}, R).
\end{equation}

Here, $\mathcal{T}=\{t_1,\dots,t_K\}$ denotes interpretable topic representations (e.g., textual descriptions or structured schemas), and $R$ denotes topic--document correspondences, potentially with auditable evidence. 

This definition is implementation-agnostic and conceptually fundamental. In this view, LLM-based topic generation naturally produces human-readable and semantically rich topic representations through prompting, going beyond static word distributions. However, the second component—explicit and testable topic–document correspondences—remains mechanistically underexplored. Although LLMs can assign topics to documents, it is not yet clear how such correspondences are internally represented and extracted. This gap motivates our study of both white-box interpretability and black-box applicability.

\section{Methods of RQ1: Attention-Informed Interpretable Structures in LLM-based TM}

Although LLMs can generate diverse and human-readable topic representations through instruction following, existing approaches do not recover the internal alignment signals that link generated topics to input documents. To address RQ1—how to recover topic--document correspondence when LLMs implicitly perform topic modeling—we propose an Attention-Informed Interpretable Structure framework that aggregates token-level attention states to fragment-level during forward inference and restores NTM-like interpretable structures in which both document--topic $\theta_d$ and topic--word $\phi_k$ distributions can be used for topic assignment and for extracting top-K representative words, respectively, without modifying the model architecture.




\subsection{Problem Setup and Experiment Basics}
In this paper, we focus on topic modeling under a practical setup: corpora with condensed semantics and documents condensed to summary-level lengths. Within this setup, the core tasks of LLM-based topic modeling consist of coherent topic generation and accurate document-topic assignment, which defines the scope of our investigation.

\paragraph{Prompt Design for White-box LLM-based TM.}
Since the primary goal of this part is to investigate the internal interpretability of the model, we adopt a relatively simple prompt to instruct the LLM to perform topic modeling over a collection of news documents. The detailed prompt template is provided in Figure~4 of the Appendix. The LLM is instructed to generate several high-level abstract topics—estimated based on the number of input documents—where each topic aggregates multiple related articles. The output is constrained to a structured JSON format and does not include intermediate reasoning traces. Our detailed prompt is shown in Figure \ref{fig:appendix-prompt1} (Appendix).


\paragraph{Token-level Attention Extraction.}
During topic generation, we record the token-level attention weights from the final Transformer layer and average them across all attention heads. Let $N_g$ denote the number of generated tokens and $N_p$ the number of prompt tokens. We then obtain an attention matrix
\begin{equation}
\mathbf{H} \in \mathbb{R}^{N_g \times N_p},
\end{equation}
where each entry is defined as
\begin{equation}
H_{t,j} = \operatorname{mean}_{\text{heads}} \bigl( \operatorname{Attn}(g_t, p_j) \bigr),
\\
\sum_j H_{t,j} = 1.
\end{equation}
Here, $g_t$ denotes the $t$-th generated token, and $p_j$ denotes the $j$-th prompt token. This matrix serves as the basis for our subsequent induction of interpretable structures.

\subsection{Attention-informed Document--Topic Distribution $\theta_d$}

We first aim to derive a topic--document relevance structure from the generation-time attention. The key intuition is that if tokens associated with a topic consistently attend to tokens within a particular document, then a semantic association exists between that topic and the document.

\paragraph{Document and Topic Fragment Alignment}
We first establish fragment-level alignment between input documents and corresponding prompt tokens, then perform fragment-level alignment between generated topics and corresponding generated tokens.
Specifically, prompt tokens are partitioned according to document boundaries, with instructional text and non-semantic tokens (e.g., punctuation and whitespace) excluded; we denote by Pd the set of prompt tokens belonging to document. Meanwhile, generated tokens are grouped into topic-specific segments, with non-semantic tokens also excluded; we denote by Gt the set of generated tokens belonging to topic t.
These procedures yield two structured partitions, 
$\{P_d\}_{d=1}^{N}$ over input documents and $\{G_t\}_{t=1}^{T}$ over generated topic segments.




\paragraph{Top-$K$ Attention Aggregation}
For each topic--document pair $(t,d)$, we collect the set of attention weights from tokens in $G_t$ to tokens in $P_d$, i.e.,
\begin{equation}
\bigl\{ H_{i,j} \mid g_i \in G_t,\; p_j \in P_d \bigr\}.
\end{equation}
Rather than averaging all such connections -- which may dilute strong semantic alignments---we select the top-$K$ attention values and compute their mean. Formally, we define the topic--document relevance score as
\begin{equation}
R_{t,d}
=
\operatorname{TopKMean}
\Bigl(
\bigl\{ H_{i,j} \mid g_i \in G_t,\; p_j \in P_d \bigr\}
\Bigr).
\end{equation}
This aggregation emphasizes concentrated attention patterns while suppressing the noise of diffuse connections. The resulting relevance matrix is denoted by
\begin{equation}
R \in \mathbb{R}^{T \times N}.
\end{equation}

\begin{figure*}[!t]
    \centering
    \resizebox{\textwidth}{!}{%
    \includegraphics[width=1.2\textwidth]{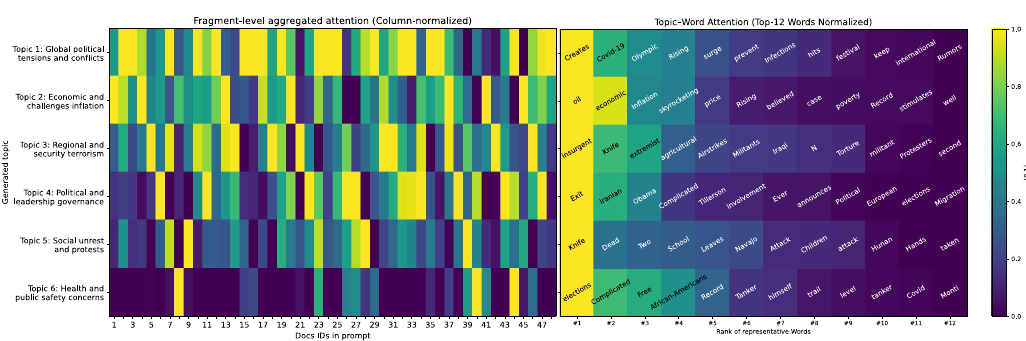}
    }
    \caption{Recovering NTM-like interpretable structures from LLM attention. The left panel shows attention-induced topic--document relevance, while the right panel shows the corresponding topic--word distributions.}
    \label{fig2}
\end{figure*}

\subsection{Attention-informed Topic--Word Distribution $\phi_k$}

Beyond the previous VAE-like distribution $\theta_d$, we can also derive a topic--word $\phi_k$ distribution to characterize the semantic composition of each topic.


\paragraph{Topic and Word Fragment Alignment.}

The alignment of the topics follows the same procedure as above, resulting in the token groups $\{G_t\}_{t=1}^{T}$.
From the document content, we extract semantic words by removing instructional text and filtering out stopwords, punctuation, and other non-semantic tokens. Let $W = \{w_1, \dots, w_M\}$ denote the resulting vocabulary.

\paragraph{Attention Aggregation.}
For each topic $t$ and word $w$, we compute the mean attention of the tokens in $G_t$ to the token positions corresponding to $w$. This gives:
\begin{equation}
A_{t,w}
=
\frac{1}{|G_t|}
\sum_{i \in G_t}
\operatorname{Attn}(g_i, w),
\end{equation}
where $\operatorname{Attn}(g_i, w)$ denotes the attention weights of the generated token $g_i$ to the prompt token(s) associated with the word $w$. Collecting these values over all topics and words yields the topic--word attention matrix.
\begin{equation}
A \in \mathbb{R}^{T \times M}.
\end{equation}

\paragraph{Debiasing and Normalization.}
To mitigate position bias and the influence of globally frequent words, 
we apply a two-step normalization procedure. 
First, for each word $w$, we subtract its mean attention from all topics:
\begin{equation}
A'_{t,w}
=
A_{t,w}
-
\frac{1}{T}\sum_{t'=1}^{T} A_{t',w}.
\end{equation}
Next, we standardize the debiased scores within each topic.
\begin{equation}
\tilde{A}_{t,w}
=
\operatorname{zscore}\!\left(A'_{t,w}\right).
\end{equation}
Finally, the standardized values are linearly rescaled to the interval $[0,1]$. 
The resulting matrix $\tilde{A}$ is treated as an interpretable topic--word distribution.

\section{Experiments of RQ1: Applications and Empirical Validation}
\subsection{Applications of Attention-informed Distribution}

\paragraph{Document-to-Topic Assignment.}
Given the relevance matrix $R$, each document is assigned to the topic with the highest relevance score:
\begin{equation}
\hat{t}(d) = \arg\max_t R_{t,d}.
\end{equation}
This yields a document--topic mapping induced purely from generation-time attention.

\paragraph{Top-$K$ Keyword Extraction.}
For each topic, we rank words according to $\tilde{A}_{t,w}$ and select the top-$K$ entries as topic keywords. If a word appears multiple times on the input, we retain its maximum score. The result is a weighted and ranked keyword list for each topic.

\subsection{Experimental Setup}
Experiments are conducted on four 80GB A100 GPUs using a locally deployed Qwen3-8B\footnote{\url{https://huggingface.co/Qwen/Qwen3-8B}} model. Although the model natively supports a context length of 32K tokens, the maximum context length is limited to 4,000 tokens due to GPU memory constraints. Consequently, we sampled 48 documents from the NYT dataset and set the number of generated topics to six.

We evaluate the validity of the attention-informed distribution indirectly through its interpretability in downstream applications. Specifically, we assess its effectiveness by examining (1) the accuracy of document-to-topic assignments derived from this distribution, and (2) human judgments of the semantic relevance between the generated keywords and their corresponding topics.

\subsection{Results and Analysis}
\paragraph{Visualization and Assignment Accuracy Analysis}

For visualization, both the topic--document and topic--word distributions are normalized to enhance contrast and plotted as Figure~\ref{fig2}. 
By normalizing the relevance scores for each document in all topics, we observe that the documents associated with different topics are relatively evenly distributed. We then asked another LLM to independently judge whether each document matches the topic with the highest relevance score. The results show that 26 out of 48 documents were correctly assigned, yielding an accuracy of 54.17\%.
In particular, 12 of the 48 documents do not belong to any of the six generated topics, implying a theoretical upper bound of approximately 75\% under this simple assignment setting. On the other hand, we can also see that the top-12 words and corresponding topics are also mainly relevant in Figure~\ref{fig2}. 
This indicates that both interpretable structures that we extracted are effective.

\paragraph{Ablation Analysis of Top-K Aggregation}
Figure~\ref{fig:topk_accuracy} shows the accuracy of the assignment under different Top-K settings, where Top-K denotes the number of largest attention weights selected from the token-level attention matrix for mean aggregation. 
Overall, accuracy increases steadily as K grows from 1 to around 100, indicating that incorporating more high-attention signals improves document--topic alignment in the low- to medium-K regime. 

Performance saturates when K reaches 100--200, where the best accuracy (0.542) is achieved. Beyond this range, further increasing K does not yield additional gains and even leads to slight fluctuations. 
This pattern suggests that aggregating a moderate number of salient attention weights stabilizes semantic estimation, while excessively large K values introduce less informative or noisy attention signals, thereby diluting the discriminative structure. The observed plateau implies diminishing returns once the major attention mass has been sufficiently captured.

\begin{figure}[t]
\centering
\begin{tikzpicture}
\begin{axis}[
    width=\columnwidth,
    height=6cm,
    xlabel={Top-K},
    ylabel={Assignment Accuracy},
    ymin=0.35,
    ymax=0.56,
    xmin=0,
    xmax=1600,
    grid=both,
    grid style={dashed,gray!30},
    tick label style={font=\small},
    label style={font=\small},
    legend style={font=\small, at={(0.5,-0.2)}, anchor=north},
]

\addplot[
    color=blue,
    mark=*,
    thick
]
coordinates {
(1,0.375)
(3,0.4375)
(5,0.4583)
(7,0.4583)
(10,0.4792)
(15,0.4792)
(20,0.5)
(25,0.5)
(30,0.4792)
(50,0.4792)
(60,0.5)
(80,0.5208)
(100,0.5417)
(150,0.5417)
(200,0.5417)
(500,0.5208)
(1000,0.5208)
(1535,0.5208)
};

\end{axis}
\end{tikzpicture}
\caption{Assignment accuracy under different Top-K token-level attention aggregation settings.}
\label{fig:topk_accuracy}
\end{figure}

\section{Methods of RQ2: Applicability of Black-Box Long-Context LLMs}
However, white-box analysis relies on direct access to hidden states and attention weights. 
In many practical scenarios, the most powerful LLMs are closed-source and accessible only through APIs, 
with their internal mechanisms remaining opaque. 
To address this limitation, we further investigate a more practical framework in black-box settings 
(RQ2: How can applicability be improved under black-box constraints, and how does performance behave—particularly with respect to long-context generalization and comprehensive comparative evaluation?).

\subsection{Reformulate as a Long-input Task.} 
For black-box LLMs with long context windows, where memory constraints do not need to be considered, we reformulate topic modeling as a structured long input task: given a corpus or an ultra-long collection of texts, 
the model generates a set of semantically coherent and mutually distinguishable topic representations. 
To compensate for the interpretable distribution loss caused by inaccessible internal state, we propose a post-generation signal compensation strategy 
which approximates latent topic mixture signals in an external memory space 
through diversified topic cue generation and hybrid retrieval mechanisms. 
This design improves the practicality of LLM-based TM in black-box scenarios. 


\subsection{Post-generation Retrieval Signal Compensation.}

\paragraph{Multi-View Topic Signal Preservation.}
To enhance post-generation alignment, we explicitly preserve multiple forms of topic representations in the prompt as retrieval cues. Each generated topic is structured to include four complementary components: (1) a concise subject label, (2) a summary describing the core semantic focus, (3) a list of representative keywords, and (4) several representative titles from source documents. These heterogeneous signals provide anchors for subsequent retrieval.

\paragraph{Composite Retrieval Signal Construction.}
Given the multi-view topic cues, we construct a composite retrieval signal using lexical, dense, and hybrid retrieval. Specifically, we adopt BGE-M3~\cite{chen-etal-2024-m3}, which unifies sparse lexical matching and dense semantic similarity within a single framework. Retrieval is performed for each type of cue (subject, summary, keywords, and representative titles), and the resulting scores are aggregated to obtain a more stable and discriminative relevance estimate.

\paragraph{Topic Assignment via Composite Relevance.}
The assignment of topic to the document is then determined based on the aggregated composite score. For each document, we compute its relevance to all candidate topics using the hybrid scoring mechanism of step 2 and assign it to the topic with the highest score. This retrieval-based formulation grounds topic assignment in both lexical and semantic evidence.

\begin{table*}[t]
\centering
\footnotesize
\setlength{\tabcolsep}{5pt}
\renewcommand{\arraystretch}{1}
\resizebox{\textwidth}{!}{%
\begin{tabular}{>{\raggedright\arraybackslash}p{1.1cm}
                >{\raggedright\arraybackslash}p{3cm}
                ccccccc}
\toprule
\multirow{2}{*}{Category} & \multirow{2}{*}{\shortstack{Topic Model or LLM \\ (Context Window)}}
& \multicolumn{2}{c}{Traditional Metrics} 
& \multicolumn{4}{c}{LLM-based Metrics} \\
\cmidrule(lr){3-4} \cmidrule(lr){5-8}
& & NPMI & Diversity & Coherence & Conciseness & Informativeness & \shortstack{Assignment Accuracy (\%)} \\
\midrule

\multirow{6}{*}{NTMs}
& ETM        & 0.1758 & 0.9200 & 2.98 & 2.54 & 2.86 & 19.6 \\
& DecTM      & 0.5542 & 0.9880 & 2.86 & 2.90 & 3.42 & 23.6 \\
& TSCTM      & 0.5364 & 0.9720 & 2.56 & 2.72 & 2.90 & 20.4 \\
& CombinedTM & 0.4853 & 0.9480 & 3.04 & 2.88 & 3.54 & 23.9 \\
& NSTM       & 0.1509 & 0.7440 & 2.80 & 2.58 & 2.84 & 19.2 \\
& ECRTM      & \textbf{0.6110} & \textbf{0.9960} & 3.16 & 2.98 & 3.48 & 20.4 \\
\midrule

\multirow{2}{*}{Others}
& BERTopic   & 0.2380 & 0.7605 & 2.97 & 2.06 & 3.08 & 31.4 \\
& FASTopic   & 0.2831 & 0.4973 & 2.44 & 2.38 & 2.75 & 24.8 \\
\midrule

\multirow{6}{*}{LLMs}
& Deepseek-v3.1 (128k)   & 0.1417 & 0.9439 & 3.46 & 3.53 & 3.93 & 78.3 \\
& Qwen3.5 (262k)        & 0.1113 & 0.9667 & 4.50 & 3.83 & 4.16 & 86.2 \\
& Llama-4 (1M)     & 0.1403 & 0.8246 & 3.90 & 4.00 & 3.90 & 69.7 \\
& GPT-5.3 (400k)        & 0.1221 & 0.9388 & \textbf{5.00} & \textbf{4.93} & 4.07 & \textbf{94.4} \\
& Claude-sonnet-4 (200k)     & 0.1886 & 0.9845 & \textbf{5.00} & 4.72 & \textbf{4.27} & 91.4 \\
& Gemini-3.1 (1M)    & 0.1634 & 0.9321 & 4.10 & 3.28 & 4.16 & 93.7 \\
\bottomrule
\end{tabular}
}

\caption{Comparison of long-context LLM-based topic modeling with other topic models.}
\label{tab:main_results}
\end{table*}

\section{Experiments of RQ2: Capability Evaluation of Long-Context and Black-Box LLMs}
\subsection{Experimental Setup}
\paragraph{Datasets and Baselines.}
We selected LLMs with the longest context windows (including both white-box and black-box models) from various vendors to evaluate their topic modeling capabilities, such as Gemini and Claude. We conducted an evaluation on the New York Times dataset, which covers various domains including politics, business, and culture, and contains 100,054 documents with around 30--50 words each. Within the allowed context window, we sample as many documents as possible and feed them to the LLM in a single prompt to generate topics. The detailed prompt template is provided in Figure~\ref{fig:appendix-prompt2} of the appendix.
We compare our approach with several NTMs baselines from the \textbf{TopMost} toolkit~\cite{wu-etal-2024-towards-topmost}, including ETM, DecTM, CombinedTM, ECRTM, NSTM, and TSCTM.
These models cover major lines of recent advances in neural topic modeling, with detailed descriptions provided in the Appendix~\ref{NTM}.
In addition, we adopt BERTopic~\cite{grootendorstBERTopicNeuralTopic2022} and FASTopic~\cite{wu2024fastopic} as additional baselines to extend our comparison beyond conventional neural topic models.

\paragraph{Evaluation Metrics and Details.}
To conduct a thorough comparison between long-context LLMs and classic TM methods, we design a comprehensive evaluation framework consisting of traditional statistical metrics and LLM-as-a-judge qualitative metrics.
Traditional statistical metrics comprise NPMI (measurement of semantic consistency of top words) and Topic Diversity (defined as the ratio of unique words among top-$k$ words across topics).
LLM-based metrics include subjective topic scoring in three dimensions: coherence, conciseness, and informativeness~\cite{10.1145/3711896.3737219}, as well as the accuracy of document-topic assignment judged by an external LLM~\cite{tan2025bridging}.





\subsection{Results and Analysis}
The evaluation Results of various NTMs and our LLM-based approach are summarized in Table~\ref{tab:main_results}.


\paragraph{Analysis of Topic Representation Capabilities Across Different Methods}
Topic examples from all models are provided in the appendix. In terms of traditional metrics, the ECRTM performs best, confirming its status as one of the most advanced NTMs. Regarding coherence, LLMs show lower NPMI scores—this can be attributed to their strong abstraction and reasoning abilities, as NPMI mainly captures literal overlaps between topics and source documents, rather than high-level semantic quality. Regarding diversity, most models achieve a topic diversity of 0.9 or higher, except BERTopic and FASTopic. This means that most models can achieve good topic differentiation.

However, in LLM-based readability evaluation, LLMs outperform others in coherence, conciseness, and informativeness, demonstrating the superiority of the LLM-based topic modeling paradigm over the traditional method in topic representation. Notably, with our post-generation signal compensation mechanism, LLMs achieve much higher topic-document assignment accuracy, highlighting the promise of black-box LLMs for topic modeling.

\paragraph{Analysis of Long-Context Processing and Topic Generation Capabilities of LLMs}

We further investigate the long-context and topic generation capabilities of various LLMs. Among all evaluated models, Claude Sonnet 4 achieves leading performance across most metrics, indicating that larger context windows and enhanced model capacity contribute to better topic modeling results. However, fine-grained experimental results in the Appendix Table \ref{tab:additional_eval_results} illustrate that context length alone cannot account for performance gaps among LLMs. Larger context windows do not guaranty improved topic relevance or diversity, nor does broader input coverage lead to more reliable document-topic assignments. Instead, different LLMs show distinct trade-offs in frequent-topic preference, positional robustness, grounding, and redundancy, which require comprehensive evaluations beyond conventional quality metrics. 

In summary, even without leading in all metrics, zero-shot LLMs are comparable or superior to strong NTM baselines in the quality of topic and assignment accuracy, while also featuring flexible representation formats, ease of use, and support for multimodal and multilingual inputs. 
These findings empirically validate the view that, as attention-informed NTMs, LLMs are better compared to traditional NTMs.

\section{Discussion}

\subsection{On the Feasibility of for Large Corpora}

Despite the extensive context windows of black-box LLMs, their capacity remains insufficient when processing large-scale corpora and lengthy documents. This naturally raises the critical question: can LLM-based topic modeling be comparable to traditional NTM methods in more challenging scenarios?
We argue that large corpora can be effectively decomposed via predefined high-level labels, while long documents can be condensed into short texts through summarization techniques. Accordingly, LLM-based topic modeling can be decoupled into two core stages: corpus preprocessing and topic modeling. Our work focuses on the second stage.


\subsection{On the Evaluation of LLM-based TM}

Traditional statistics-based metrics relying on literal overlaps are inadequate for LLM-driven topic modeling, especially for tasks requiring world knowledge reasoning and high-level abstraction.
Thus, a reliable evaluation requires comprehensive metrics coupled with human analysis.
Beyond critiquing the conventional NTM paradigm, we advocate improving traditional evaluation protocols with LLMs.
By formulating topic modeling as a long-input generation task, we can draw on evaluation practices from analogous generation tasks and design dedicated protocols that employ LLMs as judges~\cite{li2024llmsasjudgescomprehensivesurveyllmbased}.

\section{Conclusions}
We revisit topic modeling from an LLM-centric perspective and investigate it in both white-box and black-box settings. In the white-box setting, we demonstrate that generation-time attention can be leveraged to recover interpretable document–topic and topic–word structures. This validates the view that an LLM can serve as an attention-informed NTM. In the black-box setting, we reformulate topic modeling as a structured long-input task and enhance topic assignment via post-generation signal compensation. Experiments on the NYT corpus show that zero-shot long-context LLMs can outperform competitive baselines in both topic quality and topic assignment accuracy, suggesting that LLMs offer a practical and promising paradigm for topic modeling, particularly in the era of rapidly expanding context windows.
\section{Limitations}
\begin{itemize}
    \item Although our proposed method for extracting interpretable structures is simple and effective, it neglects the bias caused by attention dispersion due to the generation order of topics, which may lead to weaker attention for later-generated topics.
    \item Due to the limited internal accessibility of closed-source LLMs and constraints of local GPU memory, we have not evaluated the effectiveness at scale of the interpretable structure under the white-box setting to scenarios with extremely large parameters and ultra-long contexts.
    \item The black-box framework relies on external retrieval for post-generation compensation. Although effective, this approach introduces additional computational overhead.
    \item Previous work~\cite{chen-etal-2024-long,liu-etal-2024-lost} has shown that existing LLMs may not fully utilize information in long input contexts. This limitation may potentially lead to instability in LLM-based topic modeling, especially for models that are not the most recent or strongest available.

\end{itemize}

\bibliography{refs}
\appendix

\section{E1 Information About Use Of AI Assistants}
We use GPT-based assistance within the scope below:

1. Refinement of the draft: grammar, clarity, and conciseness edits in the author's written text.

2. Brainstorming: suggested taxonomy wording variants and potential limitation statements;

3. Appendix preparation: helped structure and summarize selected papers in a standardized tabular/sectioned format in the appendix; inclusion/exclusion decisions and final summaries were verified and revised by the authors.

4. Other trivialities such as proofreading typos and latex formatting.

The AI assistant was not used to fabricate results, run experiments, annotate data, perform quantitative analysis, or write related work without human verification.
All AI-generated content was verified by the authors.

\section{Related Work}
\subsection{Classic Topic Models and Neural Topic Models}
\label{NTM}

Topic modeling originated with Latent Dirichlet Allocation (LDA) \cite{blei2003latent,Griffiths2004FindingST}, which represents documents as mixtures of latent topics and models each topic as a word distribution under Dirichlet priors. While probabilistically grounded and interpretable, LDA relies on bag-of-words assumptions and simple linear priors, limiting its ability to capture rich semantic relationships and contextual information.

Neural Topic Models (NTMs) address these limitations by integrating probabilistic topic modeling with deep neural networks, typically within a variational autoencoder (VAE) framework. Representative models such as ProdLDA \cite{srivastava2017autoencoding} reformulate LDA using neural amortized inference, while subsequent extensions improve semantic coherence and flexibility through embedding integration (ETM \cite{dieng2020topic}), distribution decoupling (DecTM \cite{wu-etal-2021-discovering}), contextualized representations (CombinedTM \cite{bianchi-etal-2021-pre}, TSCTM \cite{wu-etal-2022-mitigating}), clustering regularization (ECRTM \cite{pmlr-v202-wu23c}), and optimal transport alignment (NSTM \cite{zhao2020neural}). These models retain interpretable latent topic structures while enhancing representational capacity through neural parameterization.

However, classical formulations—where documents are modeled as mixtures of latent topics and each topic as a word distribution—suffer from intrinsic limitations that remain unresolved. In particular, their reliance on bag-of-words (BoW) signals ignores word order and discourse structure, restricting the modeling of long-range dependencies. Moreover, such abstractions struggle to leverage world knowledge and reasoning for concept induction and semantic disambiguation, often leading to mixed, redundant, or uninformative topics. In addition, the community has long relied on a narrow set of automatic metrics (e.g., vocabulary-based topic coherence) to evaluate topic quality. These metrics do not always align with semantic usefulness or human judgment, and provide limited insight into why a topic is good or poor, or how it is grounded in supporting evidence from the source texts.

 \subsection{LLM-powered engineering for Topic modeling}

Compared with traditional end-to-end topic models, LLM-based topic modeling practices leverage the limited context window and instruction-following capability of large language models to decompose topic modeling into a set of orchestrated subtasks, emphasizing prompt engineering and pipeline orchestration. Empirical studies show that, with carefully designed prompts, direct prompting methods can rival or even replace conventional topic models in both topic discovery and topic assignment \cite{muLargeLanguageModels2024}. Existing work optimizes different stages of this pipeline: PromptTopic \cite{wangPromptingLargeLanguage2023} performs sentence-level topic extraction to better handle short texts; TopicGen \cite{bnaic2024_small_data_llm_topic} introduces a refinement pipeline based on generative priors to enhance thematic clarity for small datasets; prompt scheduling strategies improve coherence and diversity in large-scale short-text modeling \cite{doiTopicModelingShort2024}; and TopicGPT \cite{phamTopicGPTPromptbasedTopic2024a} focuses on generating human-aligned and interpretable topic labels. In addition, AgenTopic \cite{pariskangPariskangAgenTopic2025} incorporates LLMs into iterative refinement and model selection, while CHIME \cite{hsuCHIMELLMAssistedHierarchical2024b} and LLooM (Lam et al., 2024) extend LLM-based topic modeling to hierarchical organization and concept induction in domain-specific scenarios.

Overall, current LLM-based topic modeling typically follows a ``generation--refinement--assignment'' paradigm. However, most studies emphasize system design and engineering implementation, with limited discussion of the principled connections between LLMs and traditional neural topic models, and are often constrained to relatively small context windows. In contrast to complex system integration approaches, we focus on the most fundamental step in practical pipelines under an ideal setting---topic generation over long-form texts. From a theoretical abstraction and model-internal interpretability perspective, we conduct empirical analyses to uncover its intrinsic connections with classical neural topic modeling paradigms and to validate the paradigm proposed in this work.

\subsection{Interpretability and Long-Input Tasks in the LLM Era}
Recent advances in long-context modeling have expanded the scope of LLM evaluation toward \emph{long-input tasks}, where models must process and reason over extremely large contexts. Benchmarks such as LOFT \cite{leeCanLongContextLanguage2024a}, which scales to million-token inputs, assess end-to-end capabilities including in-corpus retrieval, retrieval-augmented generation (RAG), SQL-style reasoning, and large-scale in-context learning. LongInOutBench \cite{zhang2025lost} further targets realistic ``long-input + long-output'' scenarios that require global information integration and structured generation. These developments suggest a broader paradigm in which complex semantic aggregation over long corpora is treated as a unified input–output problem.

In contrast, most LLM-based topic modeling approaches continue to inherit conventional TM task formulations and NTM-style evaluation metrics, whose assumptions and boundaries are rarely articulated. We instead recast topic modeling as a structured long-form input–output task under the long-context LLM paradigm. 

Complementary interpretability research supports this perspective: recent work disentangles decoder-only Transformer language models (e.g., GPT-2, LLaMA), often characterized by representational superposition, and reinterprets them as \emph{superposed neural topic models}. By extracting semantically coherent word groups from individual neurons or computational pathways and evaluating them using topic coherence-style criteria, these studies establish a structural connection between language models and topic modeling. Along similar lines, FASTopic \cite{wu2024fastopic} proposes an encoding–alignment framework that leverages pretrained Transformer embeddings and dual semantic reconstruction to align documents with learnable topic and word embeddings, demonstrating that language model embeddings alone can be decoded into efficient and coherent topic representations.

\section{Additional Experimental Details and Fine-grained Evaluation for LLM-based Topic Modeling}
\label{app:additional-finegrained-eval}

\subsection{Additional Fine-grained Evaluation Metrics}
\label{app:additional-finegrained-eval}

\begin{table*}[t]
\centering
\small
\resizebox{\textwidth}{!}{%
\begin{tabular}{lccccccc}
\toprule
\textbf{Model} & \textbf{Context Limit} & \textbf{Avg. Texts / Topic} & \textbf{High-freq. Priority (Var.)} & \textbf{Input Coverage} & \textbf{Max Topics} & \textbf{Relevance} & \textbf{Diversity} \\
\midrule
Deepseek-chat-v3.1 & 128000 & 27.4 & 250.24 & 249/69/59 & 180 & 0.5299 & 0.95 \\
Qwen3.5 & 262000 & 39.44 & 5.25 & 100/89/49 & 189 & 0.2227 & 0.90 \\
Llama-4-maverick & 1050000 & 27.4 & 53.84 & 127/37/28 & 195 & 0.3272 & 0.90 \\
Claude-sonnet-4 & 200000 & 32.39 & 157.24 & 226/229/121 & 198 & 0.3084 & 0.80 \\
GPT-5.3 & 400000 & 30.0 & 50.0 & 427/87/41 & 175 & 0.3921 & 0.71 \\
Gemini-3.1-flash-lite & 1050000 & 39.28 & 12.27 & 427/324/213 & 189 & 0.3347 & 0.72 \\
\bottomrule
\end{tabular}%
}
\caption{Additional evaluation results for LLM-based topic modeling. Input coverage is reported as the number of assigned texts originating from the first 30\%, middle 40\%, and last 30\% of the input context. Higher relevance and diversity are better.}
\label{tab:additional_eval_results}
\end{table*}

In addition to the main evaluation results, we introduce a set of fine-grained metrics to better characterize properties that are specific to LLM-based topic modeling. Unlike conventional topic models, LLM-based approaches are influenced not only by topic quality itself, but also by generation constraints, including limited output space, long-context processing behavior, instruction-following capacity, and the tendency to generate semantically overlapping or weakly grounded topics. As a result, standard topic quality metrics alone are insufficient to fully capture the practical behavior of these systems.

\textbf{Average number of texts assigned per topic.} This metric measures how many documents, on average, are explicitly associated with each generated topic. Since the model cannot enumerate all possible document--topic assignments under output length constraints, this value reflects the effective coverage of topic-wise assignment within a limited generation budget. A larger value generally suggests that the model is able to retain more supporting texts for each topic in its output, although it may also indicate that the resulting topics are broader and less fine-grained.

\textbf{High-frequency-topic prioritization.} This metric is designed to evaluate whether the model tends to generate topics corresponding to the most frequent themes in the source corpus. To quantify this behavior, we examine the distribution of the number of assigned texts across generated topics and summarize it using the mean and variance. A larger mean suggests that the model more strongly favors topics covering high-frequency document clusters, while a smaller variance indicates that this preference is relatively stable across generated topics. This metric is useful for distinguishing models that prioritize dominant themes from those that better preserve long-tail topics.

\textbf{Input text neglect ratio.} This metric evaluates whether the model exhibits position bias when processing long-context inputs. Specifically, we track how often documents from different positions in the input sequence are reflected in the final output, and compare the contributions of texts appearing in the first 30\%, middle 40\%, and last 30\% of the input. The goal is to reveal whether the model over-relies on earlier input segments while underutilizing information from the middle portion, a pattern closely related to the \emph{lost-in-the-middle} phenomenon. A more balanced distribution indicates stronger positional robustness in long-context topic generation.

\textbf{Maximum number of topics under instruction constraints.} This metric measures how many distinct topics a model can generate while still satisfying the required output format and instruction constraints. To evaluate this property, we use a separate prompting setup that explicitly asks the model to produce as many valid topics as possible, rather than a fixed number of topics. The resulting value reflects the practical upper bound of topic generation capacity under structured output requirements, and thus provides a more realistic picture of the model's usable topic inventory than context length alone.

\textbf{Topic--text relevance.} This metric evaluates whether the documents assigned to a generated topic are genuinely consistent with that topic. It is intended to capture the degree of topical grounding and, conversely, the extent of hallucinated or weakly supported assignment. In practice, this can be measured by comparing the generated topic descriptors with the content of their assigned texts and assessing whether sufficient topical evidence is present. Higher relevance indicates that the assigned documents are better supported by the topic description, whereas lower relevance suggests a higher tendency toward hallucinated document--topic alignment.

\textbf{Topic redundancy.} This metric measures the extent to which generated topics are semantically repetitive and could be merged. A model may produce a large number of topics, yet the effective diversity of the topic inventory can still be low if multiple outputs describe essentially the same underlying theme. We therefore examine whether generated topic summaries contain redundant or highly overlapping semantic content. Lower redundancy corresponds to higher diversity, which is desirable for constructing a more informative, compact, and discriminative topic set.

\subsection{Additional Evaluation Results}

Overall, the additional results show that LLM-based topic modeling cannot be fully characterized by final topic quality alone. The models differ substantially in how they balance topic coverage, grounding, and diversity. In particular, generating more topics or assigning more texts to each topic does not necessarily lead to higher relevance or a less redundant topic inventory. This suggests that LLMs exhibit different generation preferences even under the same prompting setup.

The input coverage results further reveal clear differences in long-context behavior. Some models rely more heavily on earlier input segments, while others make more balanced use of the full context, indicating stronger robustness to positional bias. Notably, these differences are not well explained by context window size alone, suggesting that effective use of input evidence matters more than raw context capacity in LLM-based topic generation.

\subsection{Representative Topic Generation Outputs}
\label{app:qualitative-comparison}

To provide a clearer view of the outputs produced by different methods, we present representative topic generation examples in Tables~\ref{tab:ntm_results}, \ref{tab:bertopic_fastopic_results}, and \ref{tab:llm_results}. 

Table~\ref{tab:ntm_results} shows keyword-based topics generated by neural topic models. Table~\ref{tab:bertopic_fastopic_results} presents results from embedding-based methods (BERTopic and FASTopic). Table~\ref{tab:llm_results} illustrates structured outputs from LLM-based approaches, including summaries, keywords, and representative source titles.

These examples are intended to complement the quantitative evaluation by providing a direct view of the generated topics across different modeling paradigms.

\subsection{Prompt Templates for LLM-based Topic Generation}

Figures~\ref{fig:appendix-prompt1} and \ref{fig:appendix-prompt2} present the prompting templates used in our LLM-based topic modeling framework. The first prompt is designed for high-level topic abstraction, encouraging the model to generate a small number of broad themes that group multiple related documents. In contrast, the second prompt targets fine-grained topic generation, requiring the model to produce a larger set of more detailed topics with structured outputs, including summaries, keywords, and representative source titles. Together, these prompts enable us to control the granularity and structure of the generated topic space.

\begin{table*}[t]
\centering
\small
\begin{tabular}{c c p{0.78\textwidth}}
\toprule
\textbf{Method} & \textbf{Topic ID} & \textbf{Top Keywords} \\
\midrule
\multirow{5}{*}{ETM}
& 0 & china, north, south, korea, beijing, korean, nuclear, gun, latest, missile \\
& 1 & investigation, attorney, case, allegations, criminal, charges, justice, sexual, trial, prosecutors \\
& 2 & barbaro, time, whether, johnson, questions, question, point, important, right, moment \\
& 3 & court, russia, russian, case, ukraine, decision, supreme, judge, putin \\
& 4 & said, office, time, set, social, years, says, next, place, now \\
\midrule

\multirow{5}{*}{DecTM}
& 0 & pathogen, physician, virologist, minimal, cognitive, syndrome, cells, therapy, laboratory, laboratories \\
& 1 & janeiro, dilma, dictatorship, brazilians, rio, graft, jailed, aleksei, rousseff, brazilian \\
& 2 & afghans, qaeda, afghanistan, taliban, afghan, abdel, sisi, fattah, commanders, yemeni \\
& 3 & cuomo, albany, ethics, kickbacks, sheldon, businessmen, bribes, scandals, comptroller, assemblyman \\
& 4 & cyberattacks, enrichment, revolutionary, iranians, natanz, enriched, uranium, qassim, embargo \\
\midrule

\multirow{5}{*}{TSCTM}
& 0 & wobble, tumble, bonds, rout, stock, weathering, investors, fits, bond, recession \\
& 1 & jerusalem, netanyahu, israelis, palestinians, orthodox, israel, benjamin, macedonia, athens, jews \\
& 2 & prince, royal, queen, princess, meghan, throne, palace, windsor, royals, diana \\
& 3 & russian, ukrainian, soldiers, kyiv, ukraine, khashoggi, russia, offensive, civilians, ukrainians \\
& 4 & opera, musical, movies, productions, broadway, hbo, artists, premiere, queue, artist \\
\midrule

\multirow{5}{*}{CombinedTM}
& 0 & meghan, exciting, markle, harry, gifted, monarchy, millennial, titles, royal, singing \\
& 1 & voters, siena, surveys, electorate, polls, chart, upshot, voter, incomes, whites \\
& 2 & contagious, sars, humans, infected, wuhan, pneumonia, virus, china, symptoms, fever \\
& 3 & condo, irma, hurricane, rubble, surfside, champlain, flooding, storm, rico, puerto \\
& 4 & opera, magic, motives, queen, artist, cathedral, cup, movies, elegant, painting \\
\midrule

\multirow{5}{*}{NSTM}
& 0 & legal, presidential, justice, senator, election, committee, politics, vote, court, gun \\
& 1 & schools, medical, students, level, test, virus, vaccine, drug, risk, received \\
& 2 & death, outside, police, along, hours, taken, head, sunday, british, later \\
& 3 & inside, morning, front, showed, london, saturday, officers, fire, violence, killed \\
& 4 & companies, financial, added, program, plans, business, crisis, service, percent, based \\
\midrule

\multirow{5}{*}{ECRTM}
& 0 & epa, emissions, irs, pruitt, coal, climate, treasury, dioxide, warming, mnuchin \\
& 1 & catalonia, barcelona, catalan, brexit, european, spain, madrid, britain, referendum, rajoy \\
& 2 & duchess, queen, actress, princess, diana, awards, throne, royals, fame, prince \\
& 3 & indictment, wikileaks, stone, clintons, roger, operative, conspiracy, mueller, quote, hack \\
& 4 & protesters, hong, demonstrators, kong, tear, mainland, taiwan, extradition, confederate \\
\bottomrule
\end{tabular}
\caption{Representative topic generation results from neural topic models. Each topic is shown with its top keywords.}
\label{tab:ntm_results}
\end{table*}

\begin{table*}[t]
\centering
\small
\begin{tabular}{c c p{0.78\textwidth}}
\toprule
\textbf{Method} & \textbf{Topic ID} & \textbf{Top Keywords} \\
\midrule
\multirow{5}{*}{BERTopic}
& 0 & presidential, republican, trump, mr, his, democratic, clinton, campaign, candidates, donald \\
& 1 & afghanistan, al, taliban, afghan, islamic, president, pakistan, government \\
& 2 & north, korea, nuclear, south, korean, iran, secretary, president, kim, jong \\
& 3 & russia, russian, ukraine, vladimir, putin, president, moscow \\
& 4 & coronavirus, cases, hospitalizations, charts, deaths, maps, see, latest, county, virus \\
\midrule
\multirow{5}{*}{FASTopic}
& 0 & european, germany, chancellor, legislation, greece, angela, tax, austerity, merkel, budget \\
& 1 & culture, symbol, daily, generation, gay, ceremony, speaking, love, these, language \\
& 2 & ceremony, these, love, speaking, december, stopped, language, citing, era, spotlight \\
& 3 & era, elite, identity, something, these, gay, crack, brutal, daily, language \\
& 4 & killed, police, least, killing, died, protesters, dead, officers, protest, suicide \\
\bottomrule
\end{tabular}
\caption{Representative topic generation results from BERTopic and FASTopic. Each topic is shown with its top keywords.}
\label{tab:bertopic_fastopic_results}
\end{table*}

\begin{table*}[t]
\centering
\small
\resizebox{\textwidth}{!}{%
\begin{tabular}{p{0.07\textwidth} p{0.02\textwidth} p{0.26\textwidth} p{0.20\textwidth} p{0.33\textwidth}}
\toprule
\textbf{Method} & \textbf{ID} & \textbf{Summary} & \textbf{Keywords} & \textbf{Representative Source Titles} \\
\midrule
Deepseek-chat-v3.1 & 1 &
International conflicts and military interventions involving Middle Eastern and South Asian countries. &
Terrorism, Taliban, Islamic State in Iraq and Syria (ISIS), Defense and Military Forces, Afghanistan War (2001- ), Pakistan, Syria, Iraq &
ISIS Fears Prompt Britain to Raise Terrorism Threat Level; Terrorism Suspect Is Captured in a Raid in Afghanistan; Pakistan Releases Taliban Commander to Help Afghan Peace Efforts; 14 Million Children Suffering as Result of War in Syria and Iraq, Unicef Says; After Losses in Syria and Iraq, ISIS Moves the Goal Posts \\
Qwen3.5 & 1 &
This topic covers international conflicts, terrorism, and military violence, particularly focusing on instability in Afghanistan, Pakistan, and the Middle East. &
Taliban, Afghanistan, Pakistan, Terrorism, Militants, Bombings, Violence, ISIS, War, Extradition &
Pakistan Uses Tribal Militias in Taliban War; After U.S. Talks With Afghanistan, Hints at a Harder Line on Pakistan; In Advance of Cease-Fires, Bombings in Afghanistan Kill at Least 18; Terrorism Suspect Is Captured in a Raid in Afghanistan; Different Taliban Groups Claim Role in Afghanistan Bombing \\
Llama-4-Maverick & 1 &
Iran's political tensions and nuclear program &
Iran, Nuclear Weapons, Politics and Government, AHMADINEJAD, MAHMOUD, Khamenei, Ali &
Iran Confronts 3rd Day of Protests, With Calls for Khamenei to Quit; Ahmadinejad Sees Nuclear Energy in Iran by 2009; As Election in Iran Nears, Ahmadinejad's Critics Are Piling On; As Iran Nuclear Talks Resume, Ayatollah Criticizes U.S.; Breaking Silence, Ayatollah Says Iran Is Standing Up to West in Nuclear Talks \\
Claude-Sonnet-4 & 1 &
Political protests and demonstrations across various countries challenging government policies, election results, and social issues. &
Demonstrations, Protests, Riots, Politics, Government, Elections, Civil unrest, Opposition &
In an Unsettled Cambodia, Preparing to Confront the Government; Thai Premier Calls for Elections as Opposition Quits; Kenya Protests Turn Violent; Myanmar: Government Panel Approves Opposition Leader's Run for Parliament; Protests in Iran: Then and Now \\
GPT-5.3 & 1 &
Political protests and civil unrest reflecting resistance against governments and policies across regions. &
protests, civil unrest, demonstrations, political dissent, government crackdown, mass mobilization, social movements &
Kenya Protests Turn Violent; Protests in Iran: Then and Now; Student Protests Continue in Egypt Despite Crackdown; Anti-Government Protests in Venezuela Turn Violent \\
Gemini-3.1-flash-lite & 1 &
Global management of the COVID-19 pandemic, focusing on vaccine efficacy, infection tracking, and the socio-economic impact of lockdowns. &
COVID-19, Vaccine, Pandemic, Public Health, Lockdown, Infection, Immunization, WHO &
After India Jabs Millions, Its Covid-19 Vaccine Shows Potency; Traverse County, Minnesota Covid Case and Risk Tracker; Fayette County, Iowa Covid Case and Risk Tracker \\
\bottomrule
\end{tabular}%
}
\caption{Representative topic generation results from LLM-based methods. Each model is illustrated with one example topic, including its summary, keywords, and representative source titles.}
\label{tab:llm_results}
\end{table*}

\begin{figure*}[t]
\centering
\fbox{
\parbox{0.95\textwidth}{
\small
\textbf{Prompt Template for High-Level Topic Generation}

You are a professional text analyst. Analyze the provided news data and generate a SMALL number of HIGH-LEVEL, ABSTRACT topics. Your goal is to group many related articles into a few broader themes.

\textbf{Constraints:}
(1) Output only valid JSON. 
(2) Around {{estimate num}} topics. 
(3) Each topic must aggregate multiple articles.

\textbf{Required JSON Schema:}
[
    \{
        "Topic 1": \{
            "Summary": "...",
            "Keywords": ["...", "...", "...", "...", "..."]
        \},...
    \}
]

\textbf{Input Format:}

ID: 851 | Title: Rising Inflation Creates Unease in Middle East | Text: Inflation, caused in part by the skyrocketing price of oil, is pushing many ordinary people toward poverty even as oil money stimulates economic growth in the gulf.

ID: 3279 | Title: Discovering How Greeks Computed in 100 B.C. | Text: The Antikythera Mechanism organized the ancient Greek calendar in the cycles of the Olympiad, researchers say.

ID | Title | Text…………

}}
\caption{Prompt template used for high-level topic abstraction.}
\label{fig:appendix-prompt1}  

\end{figure*}

\begin{figure*}[t]
\centering
\fbox{%
\parbox{0.95\textwidth}{%
\small
\textbf{Prompting Setup for Topic Generation}\\
Please conduct a thematic analysis of the provided text data, generating several independent topics. Each topic should balance both generalization and granularity: it must provide an overall description while avoiding being overly vague.\\
IMPORTANT: For "Source Titles", you MUST copy exact titles from the input data. Look for lines containing "Title: [actual title]" and copy only those exact titles.\\
Output format:\\
    \{\{
        "Topic 1": \{\{\\
            "Summary": "A one-sentence concise summary capturing the core essence of the topic",\\
            "Keywords": "Keyword1", "Keyword2", "Keyword3", "Keyword4",\\
            "Source Titles": "Exact Title From Input Data", "Another Exact Title From Input", "Third Real Title"
        \}\}
    \}\},\\
    \{\{
        "Topic 2": \{\{\\
            "Summary": "A one-sentence concise summary capturing the core essence of the topic",\\
            "Keywords": "Keyword1", "Keyword2", "Keyword3", "Keyword4",\\
            "Source Titles": "Real Title From Data", "Another Real Title"
        \}\}
    \}\}\\
Specific requirements:\\
Generate AS MANY topics as possible, aiming for at least {max(topic\_estimate, 3)} topics but preferably MORE. Be comprehensive and exhaustive in your analysis. The first key in the dictionary (e.g., "Topic 1") serves as a numerical index for the topic.\\
FLEXIBLE REQUIREMENTS (QUALITY OVER QUANTITY):\\
- Each topic should include 5-12 keywords that are truly relevant and meaningful\\
- Each topic should include 3-8 source titles that actually belong to this topic\\
- DO NOT pad with irrelevant keywords or titles just to meet a number\\
- Focus on accuracy and relevance rather than hitting exact counts\\
- If a topic naturally has fewer keywords/titles, that's perfectly fine\\
Keywords Guidelines:\\
- Include 5-12 keywords per topic (flexible based on actual relevance)\\
- Keywords should be genuinely representative of the topic\\
- Avoid generic or filler keywords\\
- Each keyword should add meaningful information about the topic\\
Source Titles Guidelines:\\
- Include 3-8 titles per topic that genuinely belong to this topic\\
- **MANDATORY: Use ONLY exact titles from the input data (copy them exactly as they appear after "Title:")**\\
- **DO NOT create, modify, or invent any titles - they must exist in the input data**\\
- Select titles that best represent the topic's core themes\\
- DO NOT use generic placeholders like "Title1", "Title2"\\
- If fewer titles truly belong to a topic, include fewer titles\\
- Quality and relevance matter more than quantity\\
- **CRITICAL: Within each topic, do NOT repeat the same source title**\\
- **Each source title should appear only once within the same topic's Source Titles array**\\
- **VERIFICATION: Every title you include must be findable in the input text after "Title:"**\\
Semantic relevance: Keywords and titles within the same topic must share clear semantic relatedness, ensuring internal consistency.\\
Avoid redundancy:\\
- Minimize repeated keywords across topics to maintain distinction and reduce overlap\\
- **Within each topic, never repeat the same source title**\\
- **Each source title should appear only once per topic**\\
CRITICAL JSON FORMAT REQUIREMENTS:\\
- Output ONLY valid JSON format, no additional text before or after\\
- Use double quotes for all strings\\
- Avoid special characters in strings (use simple text only)\\
- Keep summaries and keywords concise to avoid JSON parsing issues\\
- Ensure the JSON is complete and properly closed\\
- If response is getting too long, prioritize fewer topics with complete information\\
Text contents to analyze:\{text\_content\}\\
}%
}
\caption{Prompt template used for LLM-based topic generation.}
\label{fig:appendix-prompt2}
\end{figure*}

\end{document}